\documentclass[letterpaper, 10 pt, journal, twoside]{IEEEtran}
\usepackage{graphicx} 
\usepackage{amsmath,amssymb}
\usepackage{nth}
\usepackage{color} 
\definecolor{mygreen}{RGB}{28,172,0} 
\definecolor{mylilas}{RGB}{170,55,241}
\usepackage{makecell}
\usepackage[utf8]{inputenc}
\usepackage{glossaries}
\usepackage{subcaption}
\usepackage[export]{adjustbox}
\usepackage{multirow}
\usepackage{svg}
\usepackage{wrapfig}
\usepackage[textsize=scriptsize]{todonotes}

\usepackage[linesnumbered,ruled,vlined]{algorithm2e}

\SetCommentSty{mycommfont}
\usepackage{xcolor}

\def\HiLi{\leavevmode\rlap{\hbox to \hsize{\color{yellow!50}\leaders\hrule height .8\baselineskip depth .5ex\hfill}}}

\DeclareMathOperator*{\argmin}{arg\,min}

\usepackage{lipsum}
\usepackage{float}
\makeatletter
\def\BState{\State\hskip-\ALG@thistlm}
\makeatother
\newcommand{\norm}[1] {||#1||}
\usepackage{tabularx,ragged2e,booktabs,caption}

\newcolumntype{C}[1]{>{\Centering}m{#1}}

\usepackage{flexisym}
\newcommand{\BEAS}{\begin{eqnarray*}}
\newcommand{\EEAS}{\end{eqnarray*}}
\newcommand{\BEA}{\begin{eqnarray}}
\newcommand{\EEA}{\end{eqnarray}}
\newcommand{\BEQ}{\begin{equation}}
\newcommand{\EEQ}{\end{equation}}
\newcommand{\BIT}{\begin{itemize}}
\newcommand{\EIT}{\end{itemize}}

\newcommand{\reals}{\mathbb{R}}

   {\begin{list}{}{%
    \setlength{\rightmargin}{0\linewidth}%
    \setlength{\leftmargin}{.05\linewidth}}%
    \sffamily\small
    \item[]{\setlength{\parskip}{0ex}\hrulefill\par%
    \nopagebreak{}}}%
   {{\setlength{\parskip}{-1ex}\nopagebreak\par\hrulefill} \end{list}}

\graphicspath{{media/}}
\usepackage{mydefs}

\usepackage{hyperref}

\ifCLASSINFOpdf
\else
\fi
\begin{document}
%
\title{\LARGE \bf
\methodname{} to Perform Blind Object Retrieval}
%
%
%

\author{Sheng Zhong$^{1}$, Nima Fazeli$^{1}$, and Dmitry Berenson$^{1}$%
\thanks{Manuscript received: September, 9, 2021; Revised December, 10, 2021; Accepted January, 10, 2022.}
\thanks{This paper was recommended for publication by Editor Markus Vincze upon evaluation of the Associate Editor and Reviewers' comments.
This work was supported in part by NSF grants IIS-1750489 and IIS-2113401, ONR grant N00014-21-1-2118, and the Toyota Research Institute. This article solely reflects the opinions and conclusions of its authors and not TRI or any other Toyota entity.} 
\thanks{$^{1}$Robotics Institute, University of Michigan, MI 48109, USA
        {\tt\footnotesize \{zhsh, nfz, dmitryb\}@umich.edu}}%
\thanks{For code, see \href{https://github.com/UM-ARM-Lab/stucco}{\tt\small https://github.com/UM-ARM-Lab/stucco}}%
\thanks{Digital Object Identifier (DOI): see top of this page.}
}
%
%

\markboth{IEEE Robotics and Automation Letters. Preprint Version. Accepted January, 2022}
{Zhong \MakeLowercase{\textit{et al.}}: \methodname{} to Perform Blind Object Retrieval} 

%



\maketitle

\begin{abstract}
Retrieving an object from cluttered spaces such as cupboards, refrigerators, or bins requires tracking objects with limited or no visual sensing. In these scenarios, contact feedback is necessary to estimate the pose of the objects, yet the objects are movable while their shapes and number may be unknown, making the association of contacts with objects extremely difficult. While previous work has focused on multi-target tracking, the assumptions therein prohibit using prior methods given only the contact-sensing modality. Instead, this paper proposes the method \methodname{} (\methodabv{}) that tracks the belief over contact point locations and implicit object associations using a particle filter. 
This method allows ambiguous object associations of past contacts to be revised as new information becomes available. We apply \methodabv{} to the Blind Object Retrieval problem, where a target object of known shape but unknown pose must be retrieved from clutter. Our results suggest that our method outperforms baselines in four simulation environments, and on a real robot, where contact sensing is noisy. In simulation, we achieve grasp success of at least 65\% on all environments while no baselines achieve over 5\%.
\end{abstract}


%
\IEEEpeerreviewmaketitle

\section{INTRODUCTION}
\label{sec:introduction}

\begin{figure}[t]
\includegraphics[width=\linewidth]{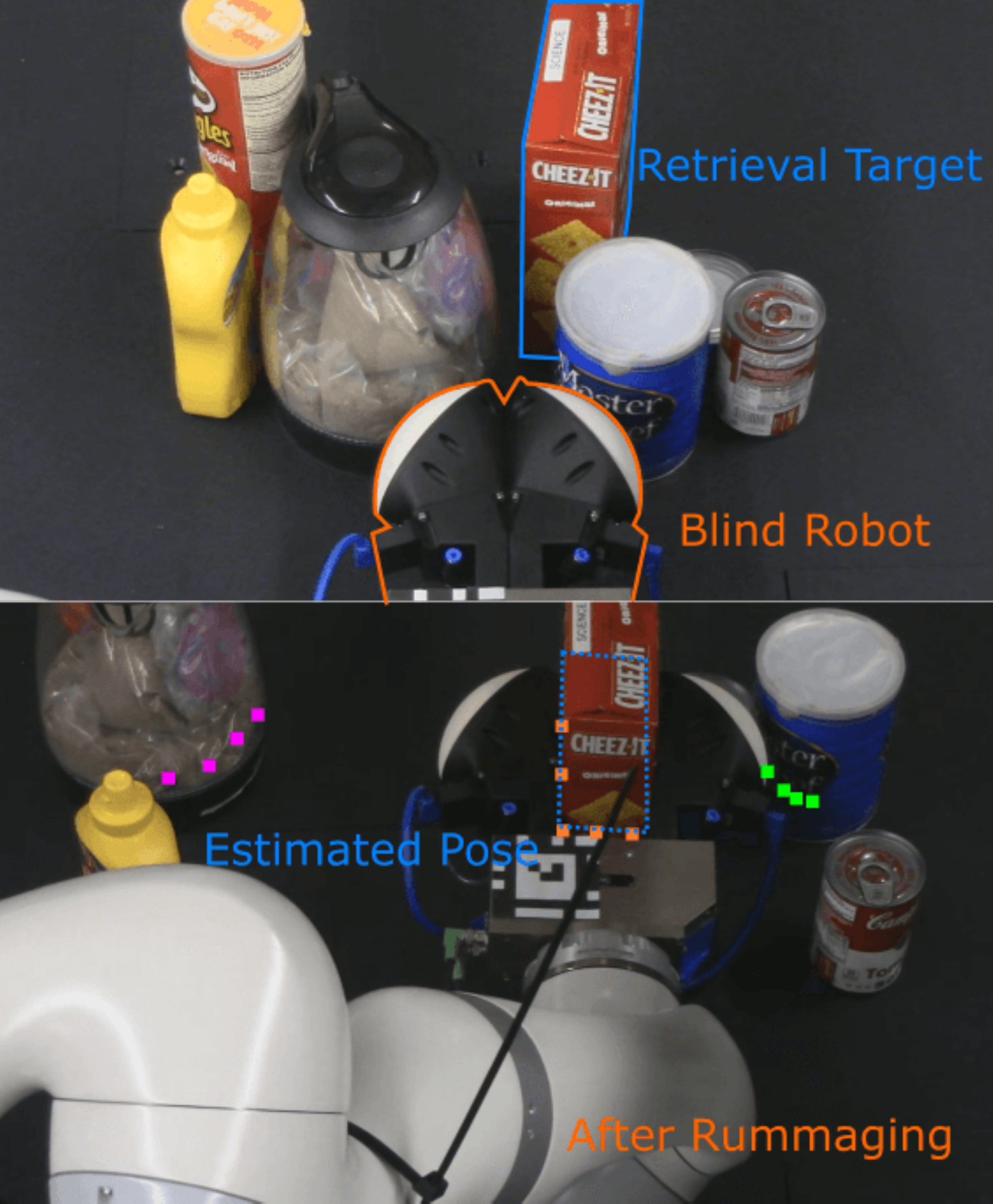}
   \caption{Initial (top) and after rummaging (bot) cluttered environment with \methodabv{} allowing us to successfully estimate the pose of the cracker box and grasp it without visual perception. The segmented tracked contact points are shown in different colors.}
  \label{fig:real}
\end{figure}

\IEEEPARstart{T}{h}is paper considers the problem of tracking objects in cluttered environments without visual feedback. Applications such as rummaging through a cupboard, refrigerator, or bin for a target object require tracking objects to estimate the pose of the target with limited or no visual sensing. In these scenarios contact feedback is necessary to estimate the poses of objects. A key difficulty is that the objects in these scenarios are movable, requiring the robot to estimate the poses of objects as they move. This is especially challenging because we do not assume we know the shapes of, or even the number of, objects in the environment \textit{a priori}. Thus when two nearby contacts are detected, it is not clear if we have contacted two objects once or one object twice. This ambiguous data association makes tracking much more difficult, as we may need to change the association of past contacts with objects when we observe new data.

Previous work in this area has focused either on single target tracking from contact \cite{koval2015pose} or on visual tracking of objects~\cite{betke2007tracking},~\cite{schwarz2018rgb}. Work on tracking multiple targets with lidar/sonar/visual data is relevant, but relies on receiving long-range information at high frequency to be effective, which is not the case for contact. To our knowledge, this is the first paper to address the problem of tracking multiple objects using only contact feedback.

The key insight that allows us to tackle this problem is that we can efficiently propagate a belief over contact points without explicit object assignments. We can then sample from that belief to generate hypotheses of contact points and associations. 
We term this approach ``soft tracking" to emphasize its difference with tracking explicit ``hard" associations.

Given a model of the pushing dynamics (which can be very simplistic) and an existing method for localizing contact on the surface of the robot~\cite{manuelli2016localizing} (contact isolation), our method, which we call \methodname{} (\methodabv{}), tracks the belief over contact point locations and implicit associations using a particle filter. We propagate the belief by sampling whether each contact point moved with probability inversely proportional to its distance to the latest contact point, then updating the particles to enforce that contact points could not have occurred inside the robot. The best estimate of contact points and a hard association of them to objects, useful for downstream tasks, can be extracted from the belief through our segmentation process. 






To show the utility of \methodabv{}, we demonstrate how it can be used to solve the Blind Object Retrieval (BOR) problem, where a target object of known shape must be retrieved from a planar cluttered environment. We evaluate our method and baselines on both simulated and real (Fig.~\ref{fig:real}) instances of this type of problem and find that our method achieves at least 65\% grasp success on all environments while no baseline achieves more than 5\% grasp success on all of them.

\section{RELATED WORK}
\label{sec:related work}
When we know the number of objects in the environment and the mapping between sensing and object is unambiguous, single target tracking methods can be used, such as ones from~\cite{wu2013online} when vision is available, or the Manifold Particle Filter~\cite{koval2015pose} when contact feedback is available. For single isolated objects, pose and shape estimation has been demonstrated using tactile feedback \cite{yu2018realtime,suresh2020tactile}. Here, we focus on the much more difficult problem when data association is ambiguous and there are an unknown number of objects.

For this problem, computer vision methods have traditionally been used. The relevant problem is termed Multiple Object Tracking, with~\cite{luo2020multiple} providing a comprehensive survey of modern methods. In cases where vision is available, its information density makes it attractive as the primary method for object detection and tracking. Our method could be used in conjunction to resolve ambiguities and provide information around occlusions. Indeed, often the robot will occlude the target as it approaches for manipulation. 

Outside of computer vision, Multiple Target Tracking is a more common term to refer to the problem and is associated with methods that are agnostic to the information source~\cite{stone2013bayesian}. Classically, Multiple Hypothesis Tracking (MHT)~\cite{blackman2004multiple} propagates hypotheses on associations of observations (contacts) to specific targets (objects). A relaxation of allowing association probabilities, instead of fixed associations, is Joint Probabilistic Data Association (JPDA)~\cite{fortmann1983sonar}. While there are ways to limit the combinatorial number of hypotheses to make these methods tractable, in the context of unknown object shapes, the explicit association of contact to objects is difficult. Often only much later do we have sufficient data to discriminate previous associations, so many hypotheses must be kept.

An alternative to explicitly considering associations is propagating the intensity (first-order moment) of the posterior on the number of targets and their states. This class of methods is called intensity filters~\cite{stone2013bayesian}, with the Probability Hypothesis Density (PHD) filter~\cite{vo2006gaussian} being a notable special case. 
We compare against an implementation of the PHD filter as a baseline. All these methods were designed with dense information sources in mind (radar, sonar, or cameras), and their assumptions are problematic in the context of blind manipulation. Most significantly, their observation models assume each target generates an observation at each step with some state-independent probability. This is clearly not the case for contact, since we can only observe contact from objects close to the robot. Our method takes inspiration from intensity filters and propagates a belief without explicit associations that exploits the local nature of contact.

Several methods have been proposed for manipulation in cluttered environments. An RGB-D approach~\cite{schwarz2018rgb} demonstrated success in visually segmenting then retrieving objects in clutter. However, the environments they showed allow immediate segmentation of the target object without needing to rummage; additionally, the objects were often well separated from each other. A haptic approach with whole-arm tactile sensing was demonstrated in~\cite{jain2013reaching} to successfully reach in clutter. In contrast to their focus on robot control for navigation, allowing them to push movable objects out of the way, we focus on perceiving the objects themselves for downstream manipulation tasks. Similar to~\cite{jain2013reaching}, our approach benefits from making numerous contacts, as each contact gives us information. While we do not have accurate localization of contact points from whole-arm tactile feedback (which is limited to very few current robots) we are able to perform our tasks using only the estimate of the external wrench at the end-effector.

\section{PROBLEM STATEMENT}
Let $\x \in \poseset$
denote the robot end-effector pose. We are given a trajectory $\x_0,...,\x_{\nsteps}$, during which the robot has made contacts with some objects. We assume that the robot is the only agent in the environment, so objects only move in direct or indirect contact with the robot. Additionally we assume that the robot moves rigidly with no compliance, the robot's geometry is known, the clutter is rigid, and that we are given a dynamics model of how objects transform for some robot motion. Our objective is to track the contact points such that they stay close to object surfaces and are segmented corresponding to the objects they belong to.

Concretely, we define \emph{contact error} (CE) on a contact point to be the smallest euclidean distance from the tracked point to any object surface. The contact error on the trajectory is the average over all contact points. Additionally, we evaluate the segmentation quality using the Fowlkes-Mallows index (FMI)~\cite{fowlkes1983method}, which approaches 0 for random assignments (with increasing number of points) and 1 for perfect assignments.

\section{METHOD}
\label{sec:methods}

\begin{figure*}[t]
\includegraphics[width=\linewidth]{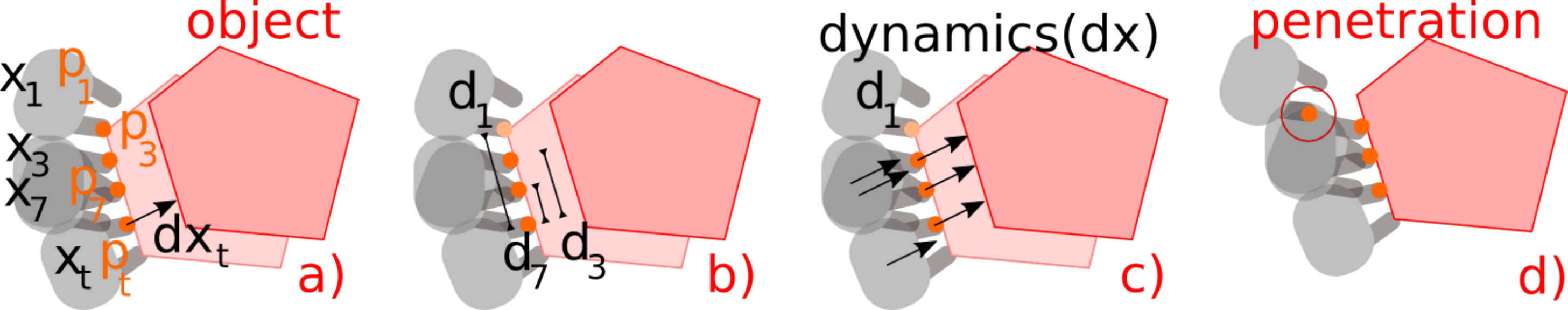}
  \caption{Prediction (a-c) and update (d) step of one particle. (a) Initial observation of the latest contact point, end-effector pose, and change in end-effector pose. (b) Compute connection probability based on distance to $\p_t$ and sample that connection; point 1 was not connected. (c) Apply dynamics to all connected points. (d) Update step assigns this particle low probability due to $\p_1$ penetrating $\x_3$ and $\x_7$.} 
  \label{fig:method}
\end{figure*}

At a high level, our approach enables downstream tasks such as object tracking and retrieval for robots ``rummaging'' in environments with only tactile feedback. To this end, our approach takes as input the robot trajectory $\x_0,...,\x_{\nsteps}$ and a set of contact points (each one denoted $\p \in \reals^3$) detected during motion. The output is the tracked set of contact points segmented based on object motions. Our method is composed of three elements: contact detection and isolation, soft tracking, and contact point segmentation, of which contact detection and isolation uses prior work while the rest are our contributions. In the following, we provide the details of each component.

\subsection{Contact Detection and Isolation}\label{sec:preliminaries}

To detect contact, we utilize the momentum observer \cite{de2005sensorless}. This observer estimates the external wrench applied to the robot ($\gamma$) using the robot's joint torques and dynamics model. We detect contact if a specified threshold $\residualcontactthreshold$ is exceeded, similar to~\cite{manuelli2016localizing,haddadin2017robot}:
\begin{equation}
    \residual^T \residualprecision \residual > \residualcontactthreshold
\label{eq:contact detection}
\end{equation}
where $\residualprecision$ is the precision matrix of the residual, measured by executing random actions in free space.

Once detected, we localize contact on the robot's surface using the Contact Particle Filter (CPF) \cite{manuelli2016localizing}. This filter iteratively solves for the contact location on the robot's surface assuming a point contact that can transmit forces but no torques -- commonly referred to as the Hard Finger approximation \cite{prattichizzo2016grasping}. We note that the remainder of our method does not depend on the details of the contact detection and isolation algorithm. As such, advances in this area can be used to extend the functionality of our approach.

\begin{algorithm}[h!]
\DontPrintSemicolon
\SetKwInput{Given}{Given}
\SetKwInput{Hyperparameters}{Hyperparameters}

\Given{
$\numparticles$ number of particles,
$\length$ characteristic length,
$\penetrationtol$ penetration length,
$\pxdist$ point to robot penetration,
$\pxdynamics$ object dynamics for change in end-effector pose}
$P_{1..\numparticles} \leftarrow \{\}$ \tcp{particles}
$t \leftarrow 0$\;
\While{robot is in execution}{
    robot executes action $\us_t$\;
    $t \leftarrow t + 1$\;
    observe $\x_t$ and $\dobj_t$, change in end-effector pose while in contact\;
    \uIf{in contact from Eq.~\ref{eq:contact detection}} {
        $\p_t \leftarrow $ get latest contact point\;
        \For{$\particleindex \leftarrow 1$ \KwTo $\numparticles$}{
            $P_\particleindex \leftarrow P_\particleindex \cup \{(\p_t, \x_t)\}$\;
            $d_{0..t} \leftarrow \norm{\p_{\particleindex,0..t} - \p_t}_2$\;\label{line:predstart}
            $p_{connect,0..t} = e^{-d_{0..t}^2/\length}$\; \label{line:connection}
            $p_{sample,0..t} \sim \mathcal{U}(0,1) $ indep.\;
            $adj \leftarrow p_{sample,0..t} < p_{connect,0..t}$\;
            \tcp{predict step}
            $P_{\particleindex, adj} \leftarrow \pxdynamics(P_{\particleindex, adj}, \dobj_t$)\; \label{line:dynamics}
            \tcp{update step}
            $\epsilon \leftarrow \sum_{i=1}^{|P_\particleindex|} \sum_{j=1}^{|P_\particleindex|} \pxdist(\x_{\particleindex,i},\p_{\particleindex,j})$\;\label{line:contact pen}
            $p_{obs,\particleindex} \leftarrow e^{-\epsilon^2 / \penetrationtol}$;\label{line:obs no contact}
        }
    }
    \Else{
        \For{$\particleindex \leftarrow 1$ \KwTo $\numparticles$}{
            $\epsilon \leftarrow  \sum_{j=1}^{|P_\particleindex|} \pxdist(\x_t,\p_{\particleindex,j})$\;\label{line:no contact pen}
            $p_{obs,\particleindex} \leftarrow e^{-\epsilon^2 / \penetrationtol}$;\label{line:obs contact}
        }
    }
    \texttt{ImportanceResample}($P,p_{obs}$)\label{line:resample}\;
    \replacebad($P$, \pxdist)
}
\caption{\methodname{}}
\label{alg:tracking}
\end{algorithm}

\subsection{Soft Tracking}

\methodabv{} maintains a belief over the positions of all contact points.
One possibility is to track each contact point independently (e.g. a Kalman Filter to estimate each contact point's position); however, this approach ignores the dependence between contact points that stems from the connectivity between points that belong to the same object. To utilize this basic assumption and represent the belief, we use a particle filter where each particle represents the set of all contact point positions and associated end-effector poses for those contacts. For convenience, we refer to the pair $(\p, \x)$
as a point. Alg.~\ref{alg:tracking} shows how we propagate this belief while Fig.~\ref{fig:method} depicts one step of our method for a single particle.

Our algorithm is structured in alternating prediction and update steps typical of Bayesian filters. Each particle is propagated independently, thus for simplicity we describe the process in terms of a single particle. However, in practice the process can be parallelized across particles and points.

Our method does not explicitly track point to object associations, like the iFilter from~\cite{stone2013bayesian} and PHD filter from~\cite{vo2006gaussian}. Instead, at each step we sample associations to predict motion, the source of the ``soft tracking" name.
Incorrect associations are propagated forward, resulting in low likelihood for the particle when the update step detects inconsistencies arising from some past mistake.

In detail, the prediction step (lines~\ref{line:predstart} to~\ref{line:dynamics} in Alg.~\ref{alg:tracking}) estimates how each contact point moves for an observed change in robot end-effector pose $d\x$. Since we assume that the robot is the only agent in the environment, we only predict motion when in contact. Contact points belong to objects; however, the likelihood of two contact points belonging to the same object scales inversely w.r.t. their relative distance. Line~\ref{line:connection} from Alg.~\ref{alg:tracking} encodes this using a characteristic length $\length$ parameter. To determine each contact point's adjacency (belonging to the same object) to the most recently encountered one, we randomly sample proportional to the likelihood provided by their relative distance. 
Contact points on the same object and their associated end-effector poses move together according to the given dynamics function $\pxdynamics$ on line~\ref{line:dynamics}. Thus a single contact point $\p_{n,i}$ at time $i<t$ 
would only move if it is adjacent to $\p_t$.

The update step in lines~\ref{line:obs contact} and~\ref{line:obs no contact} evaluates the likelihood of each particle in realizing the most recent observation, $p_{obs,\particleindex}$. We utilize the fact that contact can only occur on the robot surface to evaluate each particle. To this effect, we define the function $\pxdist(\x,\p)$ which outputs 0 if $\p$ is outside the robot when the end-effector is in pose $\x$, and otherwise $\min_{\p_s \in S(\x)} \norm{\p - \p_s}_2$, where $S(\x)$ is the set of points on the robot surface at $\x$. 

When in contact, the predicted movement of the contact points may result in penetration between any pair of $\x$ and $\p$. Thus in line~\ref{line:contact pen} we sum the penetration between all pairs in the particle. In contrast, in line~\ref{line:no contact pen} when out of contact, we only need to evaluate the observed $\x_t$ against all contact points since there is no predicted movement.

In both cases, the computation of $p_{obs,\particleindex}$
parallels our computation for adjacency during the prediction step, with a separate length parameter $\penetrationtol$ scaling with the expected contact isolation error (actual contact point's distance to the estimated contact point). A lower value will result in more false positives of penetration while a higher value will result in more false negatives. With $p_{obs}$, we perform the standard particle filter importance resampling (line~\ref{line:resample} of Alg.~\ref{alg:tracking}).

Even after resampling, particles may still have penetration inconsistencies. This could be due to none of the particles sampling a consistent prediction, or from errors in the contact isolation or contact dynamics. To address this, we call \replacebad{} after resampling, detailed in Alg.~\ref{alg:replace}. A point is inconsistent and discarded if its $\p$ incurs any penetration, and replaced with the closest consistent point in terms of $\p$ Euclidean distance in lines~\ref{line:get closest} and \ref{line:replace bad} of Alg.~\ref{alg:replace}.

\begin{algorithm}[t]
\DontPrintSemicolon
\SetKwInput{Given}{Given}
\SetKwInput{Hyperparameters}{Hyperparameters}

\Given{
$P_{1..\numparticles}$ particles, 
$\pxdist$ point to robot penetration}
\For{$\particleindex \leftarrow 1$ \KwTo $\numparticles$}{
    \For{$j \leftarrow 1$ \KwTo $|P_\particleindex|$}{
    \tcp{how inconsistent each point is}
    $\epsilon_{j} \leftarrow \sum_{i=1}^{|P_\particleindex|}  \pxdist(\x_{\particleindex,i},\p_{\particleindex,j})$\;
    }
    $incon \leftarrow \epsilon > 0$ \tcp{indices}
    \For{$j \in incon$}{
    $k \leftarrow \argmin_{i \in \neg incon} \norm{\p_{\particleindex,i}, \p_{\particleindex,j}}_2$\; \label{line:get closest}
    $P_{\particleindex,j} \leftarrow \{\p_{\particleindex,k}, \x_{\particleindex,k}\}$ \label{line:replace bad} \tcp{as well as weight}
    }
}
\caption{\replacebad}
\label{alg:replace}
\end{algorithm}

\subsection{Segmenting into Objects}
Many useful applications of tracking require a single estimate of the contact points as well as hard assignments to objects. To achieve this, our method selects the most likely particle (MAP) according to the particle weights (updated each step with $p_{obs}$). 
Alg.~\ref{alg:segment} details how the MAP particle is segmented into groups of contact points that are estimated to belong to the same object. 

Similar to line~\ref{line:connection} from Alg.~\ref{alg:tracking}, we compute the connection probability and compare it against a threshold $\connectionprob$ to determine if an edge between two points exists. The resulting adjacency matrix $A$ describes a graph over all the points, from which we find connected components. Each connected component is an object. Our segmentation is a form of agglomerative clustering (such as with BIRCH~\cite{zhang1996birch} or DBSCAN~\cite{schubert2017dbscan}), which is well suited for irregular and elongated shapes, such as the set of points belonging to surfaces of objects. A common weakness of these methods is combining two clusters when noise or an error creates a data point between them. Our update process mitigates this weakness when the robot's configuration overlaps with the erroneous contact point (depicted in Fig.~\ref{fig:snap}) and it is deemed inconsistent, but it remains an issue if our robot does not explore that location.


\begin{algorithm}[tb]
\DontPrintSemicolon
\SetKwInput{Given}{Given}
\SetKwInput{Hyperparameters}{Hyperparameters}

\Given{
$P_\particleindex$ a single particle, 
$\length$ characteristic length,
$\connectionprob$ probability threshold for each edge}
\For{$i \leftarrow 1$ \KwTo $|P_\particleindex|$}{
    \For{$j \leftarrow 1$ \KwTo $|P_\particleindex|$}{
    $d \leftarrow \norm{\p_{\particleindex,i} - \p_{\particleindex,j}}_2$\;
    $A_{i,j} = e^{-d^2/\length} > \connectionprob$\; \label{line:hard assignment}
    }
}
\Return connected components of adjacency matrix $A$
\caption{Segment a particle into objects}
\label{alg:segment}
\end{algorithm}

\section{EXPERIMENTS}
\label{sec:results}
In this section, we evaluate and benchmark the performance of our approach on: i) tracking and segmentation of contact points under a ``blind rummaging'' policy; and ii) a downstream task of Blind Object Retrieval (BOR) -- both in cluttered environments. To this end, we first describe our baselines. Next, we describe the robot environment and training data. 
Then, we formalize the downstream BOR task that uses contact tracking. Lastly, we quantitatively evaluate our method and baselines on BOR in simulated and real-world cluttered environments. 

For all tasks, we used the following $\pxdynamics$: 
\begin{equation}
    \pxdynamics(P_{\particleindex}, \dobj) = \{(\p + F(\dobj), \x + \dobj) | (\p,\x) \in P_{\particleindex}\}
\end{equation}
where $F$ extracts the linear translation of the pose change.
This motion model implicitly assumes that objects translate together with the robot when in contact without rotation. A more sophisticated motion model may be used if object properties such as size, shape, or pressure distribution are known \textit{a priori}; however, this is not the case in our experiments. Here, we demonstrate that our method is able to partially mitigate errors from this approximation since some of its predictions result in contact point penetration. 

To speed up our method, we implemented Alg.~\ref{alg:tracking} to process each particle in parallel. In particular, \pxdist{} was implemented as a parallel lookup of a pre-computed discretized (resolution 1mm) signed distance field of the end-effector in link frame. The transform of contact points from world to link frame was also implemented to be parallel.

For contact isolation, we used the Single-CPF from~\cite{manuelli2016localizing} which assumes each detected contact occurred at only one contact point. We note that while there are inherent ambiguities in isolating contact from externally-applied wrenches, our method is robust to these errors. On the real robot, we additionally consider contacts detected by each of the two soft-bubble sensors~\cite{kuppuswamy2020soft} (seen in Fig.~\ref{fig:real}). This distributed tactile sensing modality significantly mitigates ambiguities from using only wrench estimates. 

\subsection{Baselines}
We compare against baselines that maintain a single estimate of all $\p$, clustering on $\p$ at each step in contact, and applying the dynamics function to all points in the same cluster as $\p_t$. The baselines differ in their clustering methods, with BIRCH~\cite{zhang1996birch} and DBSCAN~\cite{schubert2017dbscan} by default not needing to specify the number of clusters. A k-means baseline was implemented that starts with a single cluster and increases the cluster number by 1 if doing so reduces the inertia (what k-means minimizes) sufficiently. Additionally, we consider a Gaussian Mixture (GM) implementation of the PHD filter~\cite{vo2006gaussian}. As introduced in Section~\ref{sec:related work}, this method propagates the intensity (first-order moment) of the posterior on the number of objects and their positions. The intensity is integrated over to extract discrete targets (objects), which we clustered the contact points to using nearest neighbours then propagated in the same way as the clustering methods.

\subsection{Training Set for Tuning}
\begin{figure*}[t]
\includegraphics[width=\linewidth]{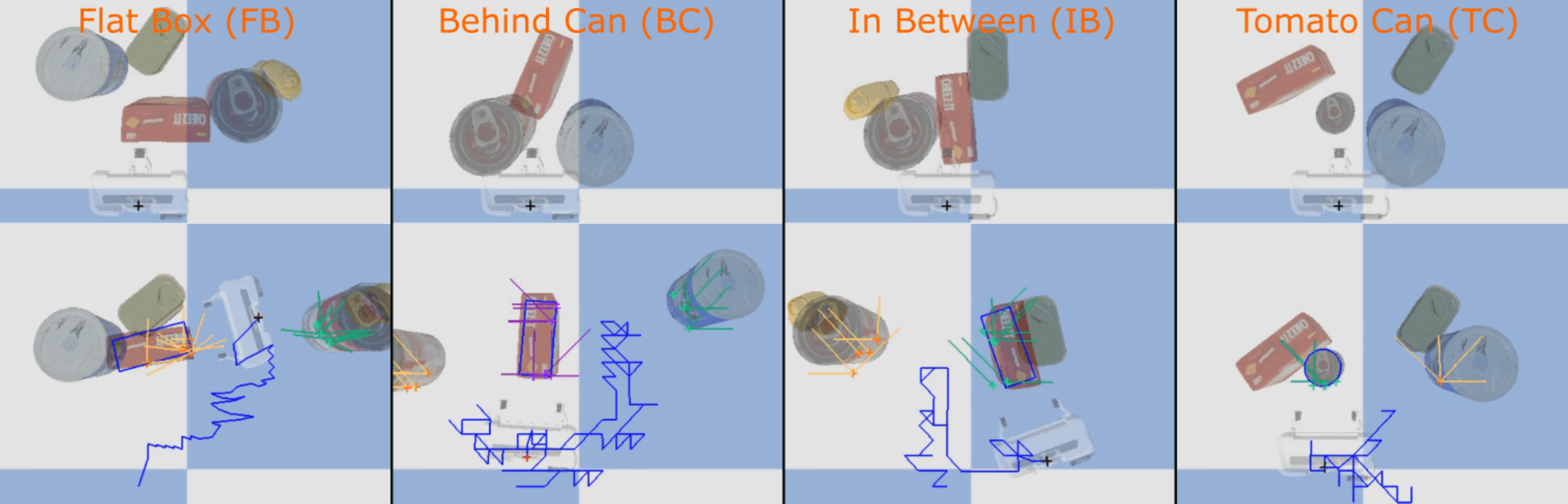}
  \caption{Simulated BOR task in 4 different environments, with each starting condition (top) and after executing actions (bottom), with the trail in blue. Overlaid is \methodabv{}'s best estimate of segmented objects, with propagated contact points as crosses and associated actions taken with a different color for each object. The pose estimate of the target object is represented by a blue outline.}
  \label{fig:bor sim}
\end{figure*}

For simulation, we use a floating Franka Emika (FE) gripper from the PANDA arm (see Fig.~\ref{fig:bor sim}) with a fixed height and constrained orientation. The gripper is simulated in PyBullet~\cite{coumans2016pybullet} and takes discrete action steps in the form of desired $dx, dy$, with a maximum per step movement of 0.03m along each dimension. Each simulation time step is 1/240s, and we moved slowly to avoid bouncing objects off the robot.
The residual $\residual$ used for contact detection and isolation here is the measured force torque on the gripper provided by the simulator.

Note that our method takes contact points as input and is not limited to planar systems. However, restricting to planar motion simplifies the data collection, contact isolation, and the downstream task of BOR.

\begin{figure}[t]
\includegraphics[width=\linewidth]{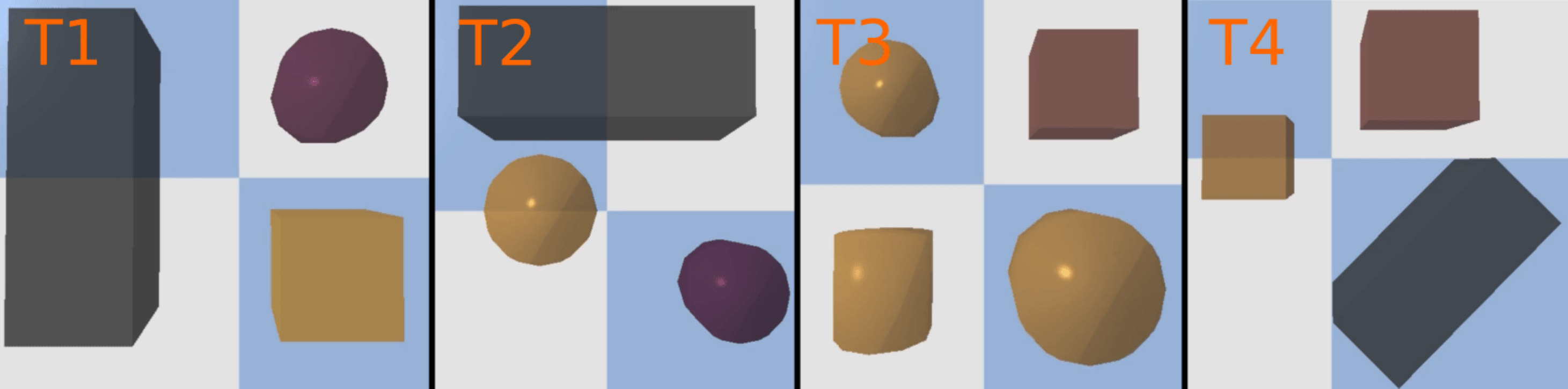}
   \caption{Training environments in simulation. Immovable walls are colored grey while the darker the movable object is, the more massive it is.}
  \label{fig:training}
\end{figure}

Our training set consists of 40 trials of randomized start and goal positions for each of the 4 environments depicted in Fig.~\ref{fig:training}. We generated trajectories using a greedy controller that entered a random walk of length 6 upon contact. Trajectories that were in contact less than 5\% of the time were discarded without replacement, yielding a total of 129 valid trajectories.

Tuning consisted of parameter sweeping to maximize median FMI and minimize median CE on the whole training set. For our method, the primary parameter to tune was the characteristic length $\length$, which was larger on the real robot to handle a large kettle. See Tab.~\ref{tab:parameters} for our tuned parameters. BIRCH was tuned to have threshold 0.08, DBSCAN was tuned to have eps 0.05 and minimum neighbourhood size of 1, k-means was tuned to need an inertia improvement of 5 times to increase the number of clusters, and the GMPHD filter was tuned to have birth probability of 0.001, spawn probability of 0, and detection probability of 0.3.

\begin{figure}[]
\centering
\includegraphics[width=0.85\linewidth]{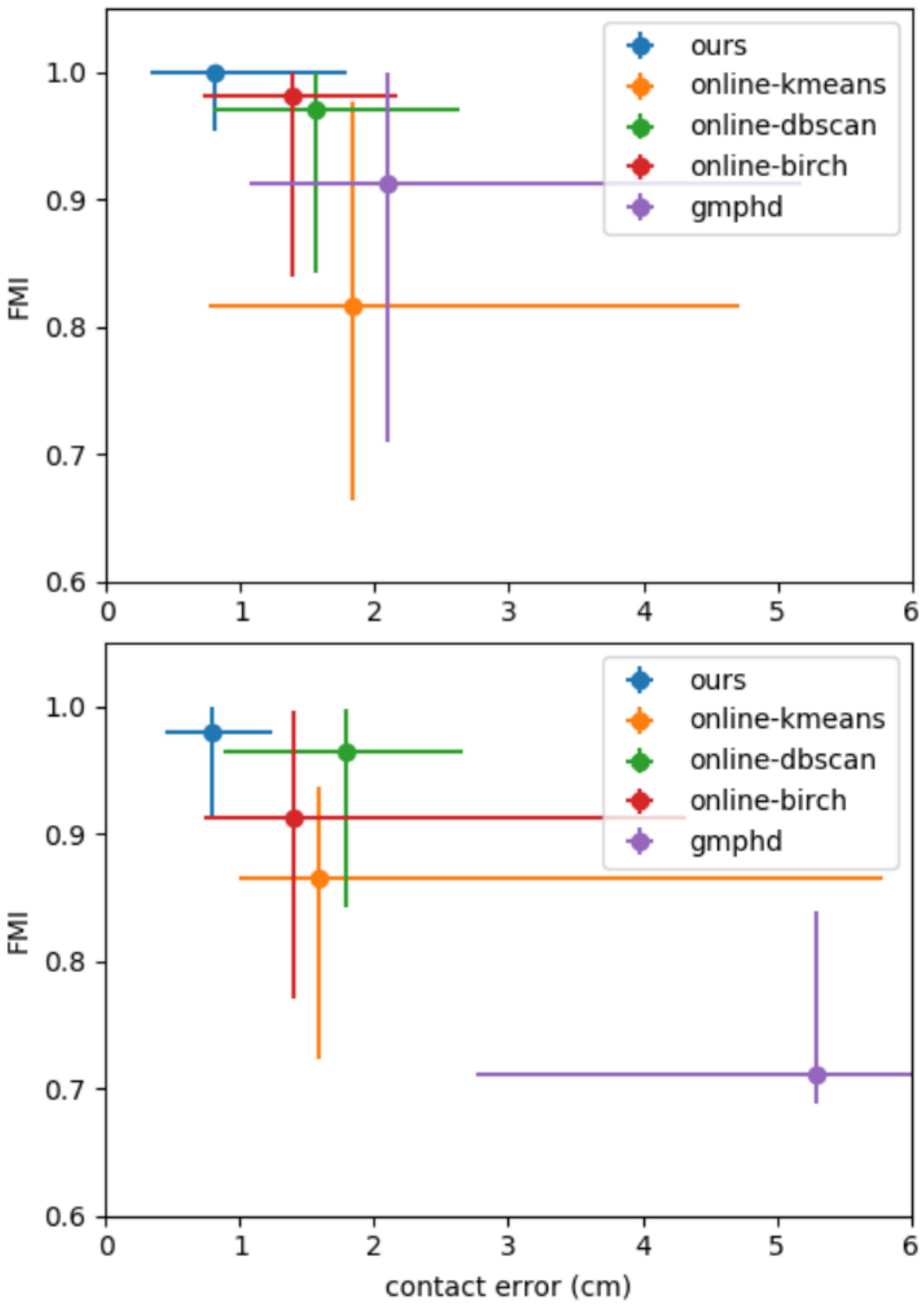}
   \caption{Tracking metrics evaluated on the training set, with the median plotted and error bars indicating 20-80$^{th}$ percentile. (top) Results for the whole data set, and (bot) results for only runs with an ambiguity score of at least 0.3. Ideal performance is an FMI of 1 and CE of 0 corresponding to points in the upper left.}
  \label{fig:training tracking}
\end{figure}

The tuned performances of all methods on the training set are shown in Fig.~\ref{fig:training tracking} (top), where our method outperforms all baselines in CE. Since we are interested in manipulation in clutter, Fig.~\ref{fig:training tracking} (bottom) shows the performance on runs that had ambiguous contact assignments. For each step, this was computed using the minimum distance from the robot to the second closest object, with an ambiguity score of 1 corresponding to a distance of 0 and a score of 0 corresponding to a distance of 0.15m or more. The bottom figure shows runs with an average ambiguity of at least 0.3. On these, our method outperforms the baselines in both FMI and CE by an even larger margin, demonstrating that our method is well suited for clutter.

\subsection{Blind Object Retrieval}

\begin{table}[tb]
\caption{\methodabv{} parameters for Blind Object Retrieval.}
\label{tab:parameters}
\centering
\begin{tabular}{|l|c|c|}
\hline
Parameter & sim & real\\ \hline
$\length$ characteristic length & 0.02 & 0.006\\
$\penetrationtol$ penetration length & 0.002 & 0.002\\
$\residualcontactthreshold$ residual threshold & 1 & 5\\
$\numparticles$ number of particles & 100 & 100\\
$\connectionprob$ connection threshold & 0.4 & 0.4 \\ \hline 
\end{tabular}
\end{table}

\begin{figure}[h]
\includegraphics[width=\linewidth]{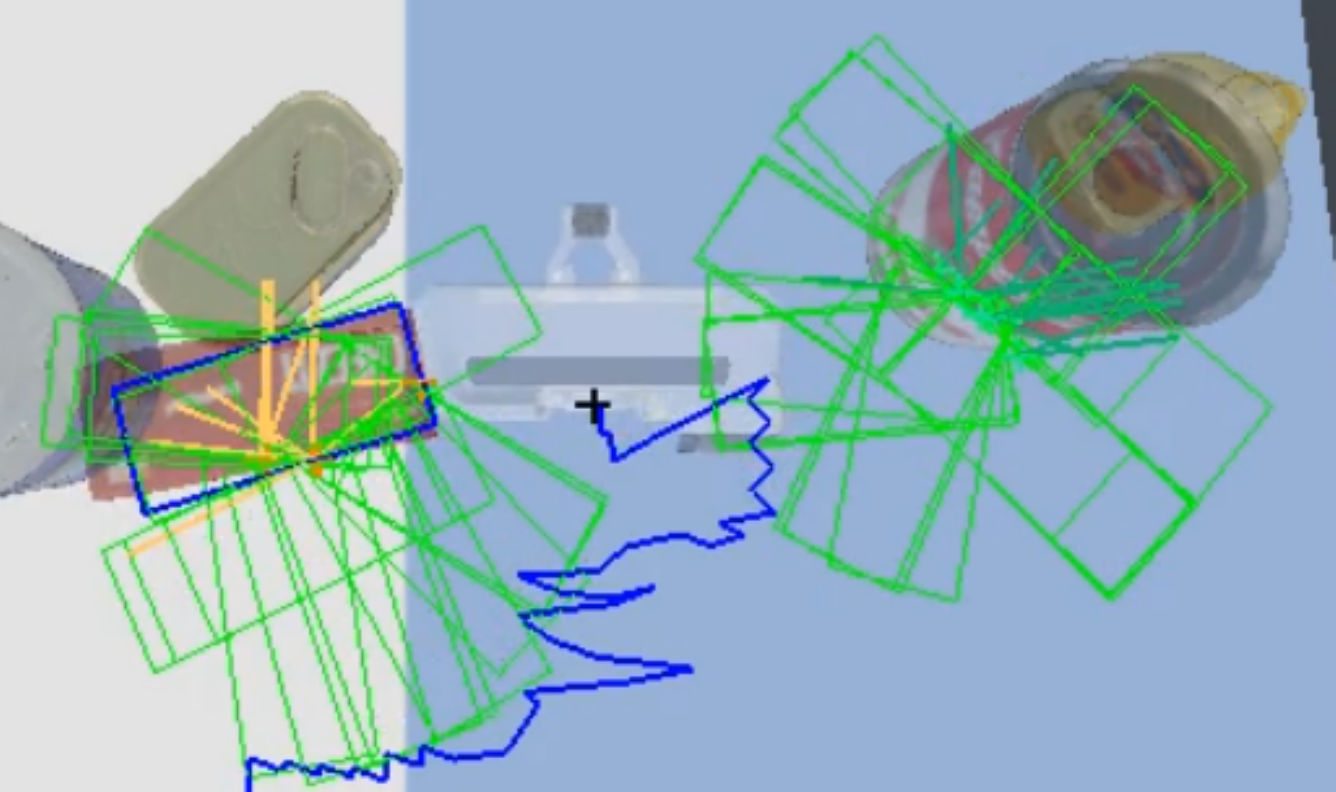}
   \caption{Iterative closest point pose estimates from 30 random starts plotted in green. ICP run on the left set of points (corresponding to the cracker box) result in lower positional variance in the estimate than the ICP results run on the set of points corresponding to the can (right).}
  \label{fig:icp}
\end{figure}

\begin{figure*}[t]
\centering
\includegraphics[width=0.329\linewidth]{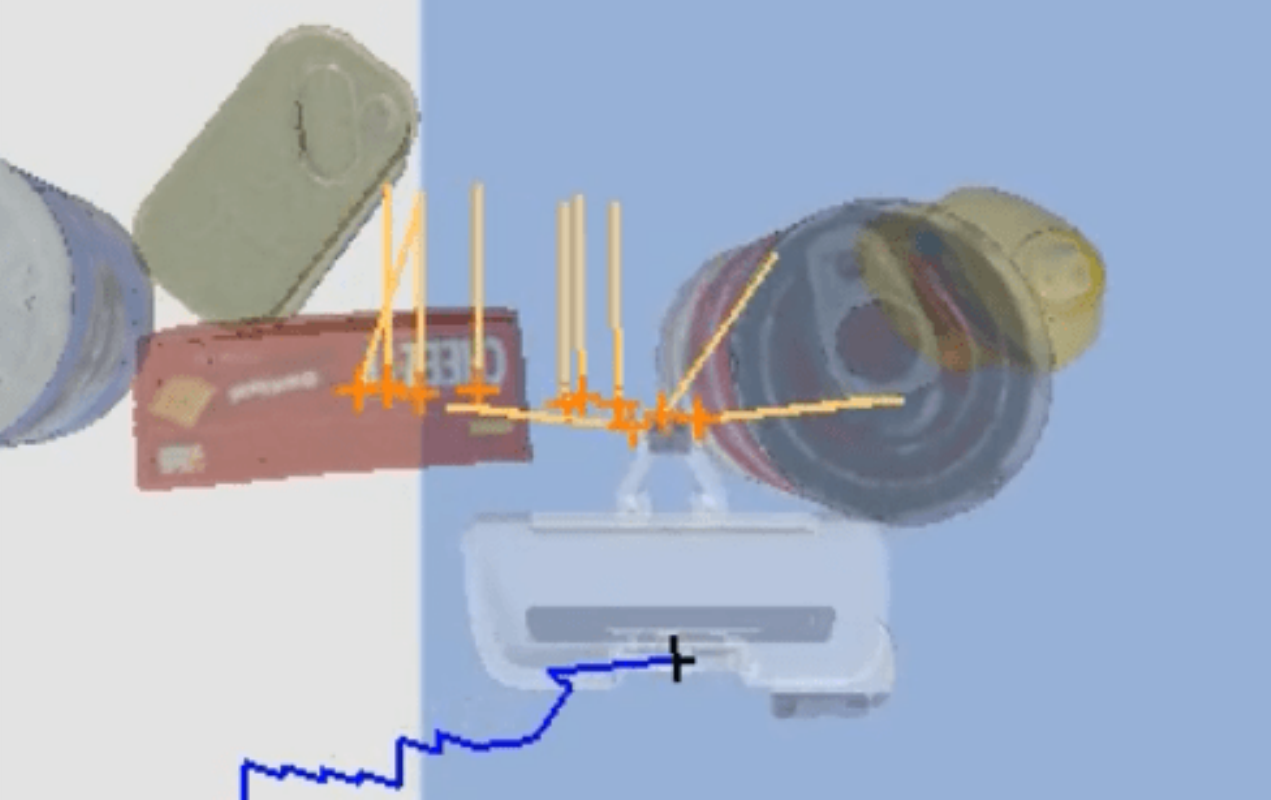}
\includegraphics[width=0.329\linewidth]{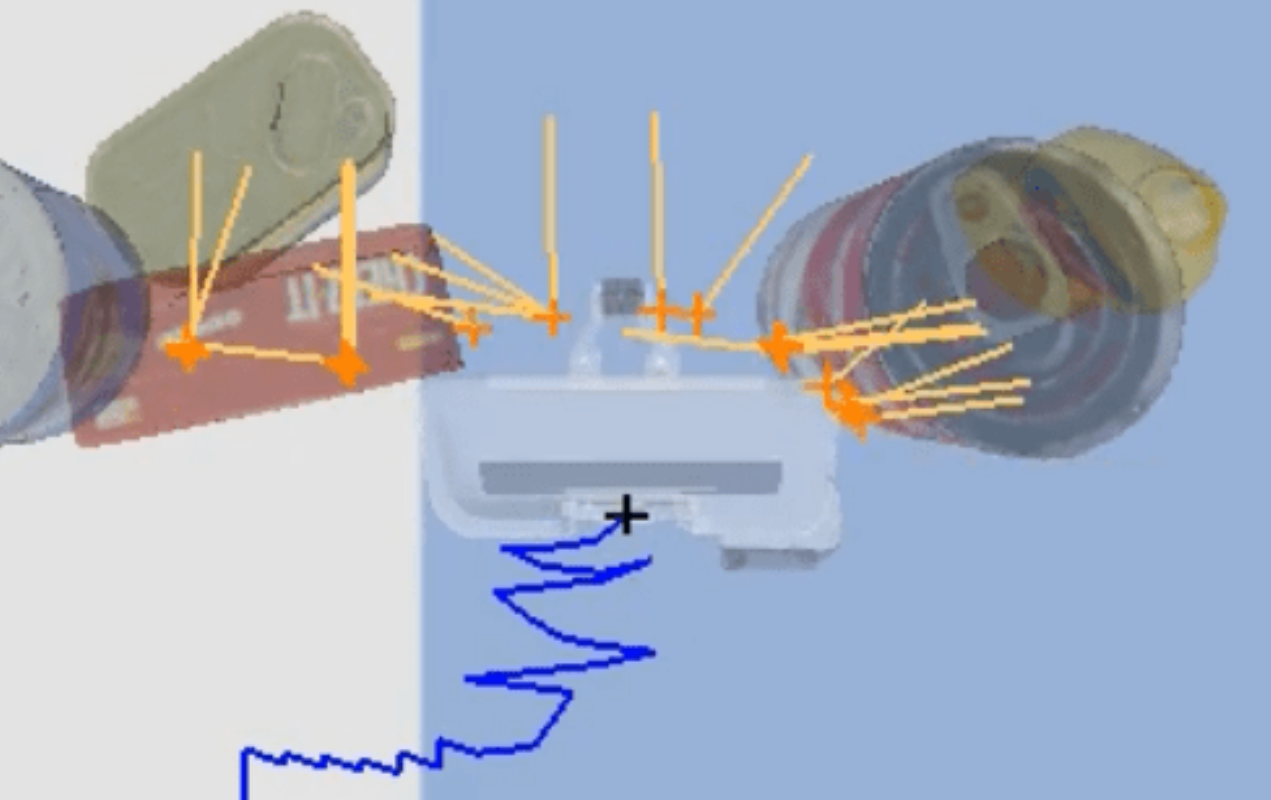}
\includegraphics[width=0.329\linewidth]{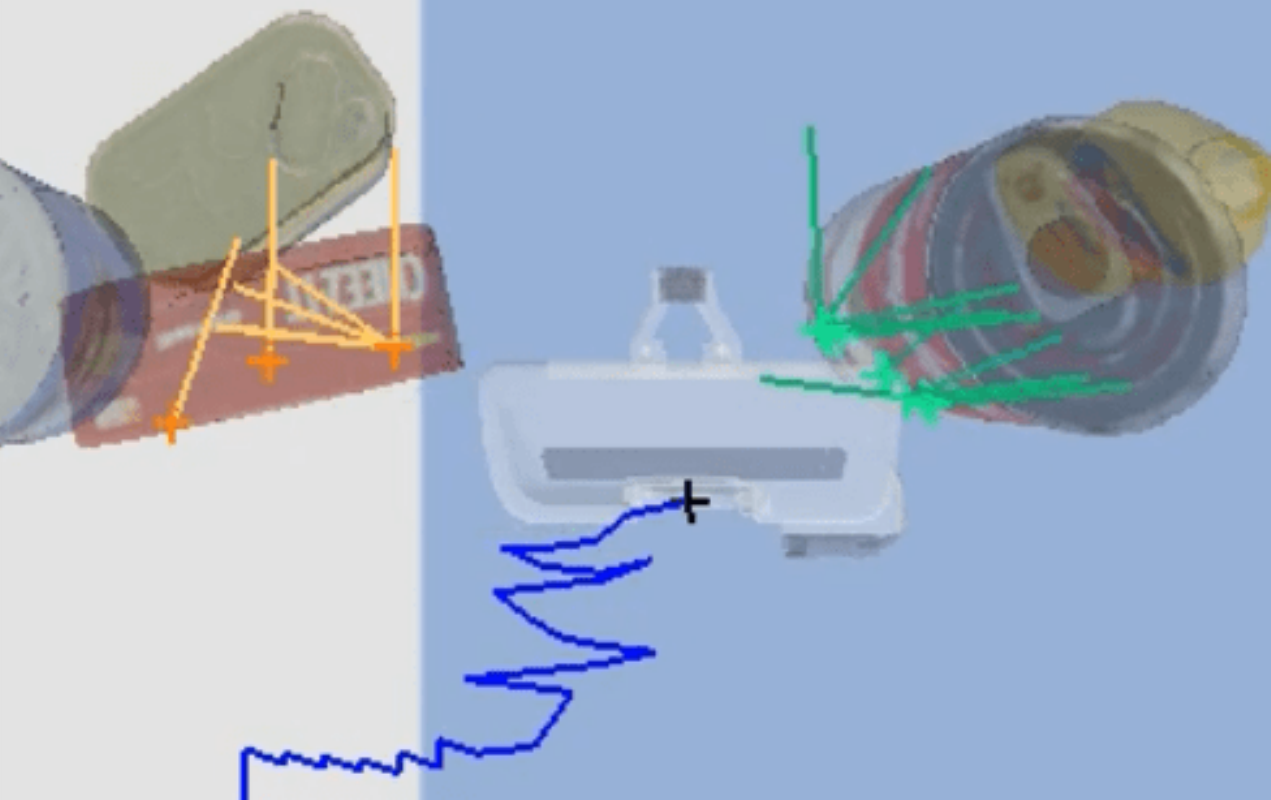}
   \caption{Steps during a BOR run with (left) initially wrong associations of contact points to objects, (mid) moving to right before entering the gap between the objects, and (right) resolving the previous ambiguity from moving through the gap and penalizing particles with points in between. Tracked contact points are in orange for the first object and green for the second one.}
  \label{fig:snap}
\end{figure*}

We present the problem of Blind Object Retrieval: pose estimation and grasping of a target object with known geometry using no visual perception. To perform this task, the robot rummages in clutter to collect contact points that it can segment into objects using our method. Using the segmented objects, the robot runs iterative closest point (ICP)~\cite{arun1987least} between the set of contact points of each object and model points sampled from the known surface of the target object. ICP is run 30 times from random initial poses and the object with the lowest variance in position estimation is selected (see Fig.~\ref{fig:icp}). An important source of variance is uncertainty in orientation due to contacts being unevenly distributed across the object surface. 
From the ICP estimates of the selected object, we further select the one that penetrates $\x_t$ the least, and on ties choose the lowest ICP matching error.

To evaluate success, we attempt a
grasp at the estimated pose after executing a given rummaging policy. Grasp success is an important metric to evaluate on for two reasons. First, it avoids the need to combine position and orientation errors. These metrics are typically combined using radius of gyrations which may not be available during run time. Second, it does not penalize small pose errors that may not be relevant to the task and can safely be ignored while penalizing those beyond a gripper dependent threshold that will always result in grasp failure.

\subsection{Blind Object Retrieval in Simulation}

In simulation, we designed 4 cluttered environments (see Fig.~\ref{fig:bor sim}) with YCB objects~\cite{calli2017yale}, with the target being the cracker box in FB, BC, and IB, and the tomato can in TC. A grasp was successful if it closed on the two long sides of the box and not on a corner, or around more than half of the tomato can. The control sequence was manually created to make contact with multiple objects that were initially close together, while making sufficient contacts to identify the target. When replayed, each action was perturbed with uniformly random noise of up to $\pm0.5$mm (compared to max action step of 30mm).

We performed 20 runs of each task (same random seed used for each baseline so they are evaluated under the same actions), with the statistical comparison of our method against baselines summarized in Tab.~\ref{tab:res}. Our method was the only one that achieved 65\% grasp success or higher on all tasks, and achieved significantly lower contact error than all baselines. 

Our method also achieved better FMI than baselines. 
The overall lower FMI scores compared to the training set seen in Fig.~\ref{fig:training tracking} attests to the difficulty of the BOR tasks. Importantly, the lowered FMI (indicating more assignment errors) for our method did not translate to increased CE. This is a key strength of our method and can be attributed to the particle update penalizing contact penetration and also the \replacebad{} function. Specifically, the action sequences often moved back and forth between two objects, eventually opening a gap and moving through it, as captured in Fig.~\ref{fig:snap}. The oscillatory motion initially left particles with contact points between the gap, but after moving through it, the update process assigned high likelihood to particles that separated the contact points to either side. Additionally, \replacebad{} replaced remaining points in between with points on a side while occasionally making an assignment error. 

The baselines have no mechanisms for correcting associations in hindsight, so the lower FMI translated to higher CE and lower grasp success. However, despite the high errors in CE, the baselines sometimes achieved grasp success due to the ICP eliminating many wrong pose estimates. 

\begin{table}[b]
\caption{Quantitative comparison of our method against baselines on 20 runs of blind object retrieval in different simulated cluttered environments and 5 runs on the real environment in Fig.~\ref{fig:real}. GS is grasp success (\%), and CE is contact error (cm). Top values per category are in bold while standard deviations are in parentheses. FB, BC, IB, and TC are depicted in Fig.~\ref{fig:bor sim}.}\label{tab:res}
\tabcolsep=0.09cm
\begin{tabular}{|ll|l|lll|l|}
\hline
\multicolumn{2}{|l|}{task} & ours        & BIRCH       & DBSCAN      & k-means     & GMPHD       \\ \hline
\multirow{3}{*}{FB}  & GS  & \textbf{70}          & 5           & 5           & 45          & 25          \\
                     & FMI & \textbf{0.69 (0.08)} & 0.62 (0.07) & 0.66 (0.10) & 0.63 (0.05) & 0.56 (0.05) \\
                     & CE  & \textbf{0.72 (0.25)} & 4.63 (0.53) & 3.52 (1.54) & 1.49 (0.40) & 3.90 (0.61) \\ \hline
\multirow{3}{*}{BC}  & GS  & \textbf{80}          & 0           & 10           & 5          & 0           \\
                     & FMI & \textbf{0.92 (0.05)} & 0.84 (0.09) & 0.89 (0.02) & 0.83 (0.04) & 0.57 (0.04) \\
                     & CE  & \textbf{1.33 (0.13)} & 6.49 (0.36) & 6.56 (0.31) & 7.04 (0.41) & 7.33 (0.44) \\ \hline
\multirow{3}{*}{IB}  & GS  & \textbf{65}          & 0          & 60          & 10          & 0           \\
                     & FMI & \textbf{0.78 (0.07)} & 0.71 (0.05) & 0.75 (0.06) & 0.47 (0.23) & 0.46 (0.06) \\
                     & CE  & \textbf{1.04 (0.33)} & 3.27 (0.36) & 2.31 (0.58) & 4.84 (1.81) & 5.62 (0.85) \\ \hline
\multirow{3}{*}{TC}  & GS  & \textbf{85}          & 0          &  0          & 25          & 20           \\
                     & FMI & \textbf{1.00 (0.00)} & 0.79 (0.07) & \textbf{1.00 (0.00)} & 0.54 (0.17) & 0.54 (0.17) \\
                     & CE  & \textbf{0.31 (0.90)} & 4.20 (0.22) & 4.14 (0.09) & 6.23 (0.72) & 8.09 (0.22) \\ \hline
Real                 & GS  & \textbf{100}         & 20          & 0           & 20          & 0           \\ \hline
\end{tabular}
\end{table}

\subsection{Real Robot Blind Object Retrieval}

We applied our method on a real 7DoF KUKA LBR iiwa arm with two soft-bubble tactile sensors~\cite{kuppuswamy2020soft} on the gripper for the BOR
task depicted in Fig.~\ref{fig:real}. Similar to simulation, we restricted our motion to be planar, with a max step of 20mm implemented using a Cartesian impedance controller. KUKA's on-board software estimated the externally applied wrench at the end-effector using the measured joint torques. To accommodate the limited sensitivity of this measurement, We filled the YCB objects to increase their mass and reduce the effect of measurement noise.

Contact isolation was performed independently by the left and right soft-bubble sensors, then by the CPF if neither of them detected contact. Each bubble had a depth camera inside that measured surface deformation. Pixels with deformations greater than 4mm were considered deformed. We then averaged all deformed pixel coordinates and projected that point to the camera frame then rigidly transformed it to produce a contact point in the world frame. Due to the deformable nature of the sensors, we adjusted $\pxdist{}$ to ignore the first 10mm of penetration.

To extend Alg.~\ref{alg:tracking} to multiple contact points per step, each contact point's distance in line~\ref{line:predstart} is measured against their closest new contact point, and dynamics in line~\ref{line:dynamics} is performed with their associated $\dobj$. Note that $\dobj$ for each new contact point may be distinct due to making contact at different times during the action.

\section{DISCUSSION}
\label{sec:discussion}
Proprioceptive and tactile driven object state-estimation is an important functionality for autonomous robotic systems in highly unstructured environments. Here, we discuss important extensions that can further generalize our method to more challenging instances of ``blind'' object state-estimation for downstream tasks such as object retrieval. These extensions include generalizing beyond point contacts and generating the rummaging policies.
\subsection{Generalizing Beyond Object Translation}
Our update step can use inconsistency in object motion to correct for errors in the dynamics function, such as assuming that objects translate without rotation. However, it is unable to handle higher dimensional pose changes such as those induced by toppling or deforming. To address this, more rich information beyond point contacts can be extracted from each contact (e.g., incipient slip from the soft-bubbles) together with more sophisticated object dynamics models.
\subsection{Generalization Beyond Single Point Contacts}
Alg.~\ref{alg:tracking} can generalize beyond single point contacts without major changes. Indeed, as shown in our real robot experiment with essentially three contact detectors, we can easily generalize to multiple contacts per step. The soft-bubble sensors provide rich contact information that we hope to exploit in future work. Advanced representations of contact patches such as meshes and non-uniform rational B-splines (NURBS)~\cite{piegl1996nurbs} could directly replace or exist alongside contact points in Alg.~\ref{alg:tracking} as long as we have efficient pairwise distance functions between them.
\subsection{Rummaging Policy}
In this paper, our method assumed a prescribed action sequence that makes sufficient contact with our target object to uniquely identify it. Future work could generate this policy using active perception. For example, after estimating the object pose using a probabilistic method (or approximations such as using multiple runs of ICP to estimate the target pose and associated uncertainty), an action sequences can be chosen to reduce uncertainty while maintaining distance from other contact point sets to minimize ambiguity.

\section{CONCLUSION}\label{sec:conclusion}
We presented \methodabv{}, a contact tracking method that maintains a belief over contact point locations to enable corrections in hindsight. The method is based on the basic assumptions that points closer together are more likely to be on the same object, and that contact only occurs on the surface of the robot. We showed that it performs well on a variety of Blind Object Retrieval tasks in clutter and demonstrated its application on a real robot.
Specifically, it is capable of handling cases where contact is initially made on different objects close together, and later correct their tracking when they are moved apart. In contrast, we showed that clustering and the PHD filter baselines struggle in these scenarios. 
Finally, the failure of baselines that maintain a single estimate of the contact points on our tasks suggests that it is beneficial to maintain a belief over them to allow corrections in hindsight. Future work will focus on representing rich contact patches and generating the rummaging policy.

\appendices
\renewcommand{\thesectiondis}[2]{\Alph{section}:}


\ifCLASSOPTIONcaptionsoff
  \newpage
\fi



%

\bibliographystyle{IEEEtran}
\bibliography{myrefs}




\end{document}